\documentclass[10pt,twocolumn,letterpaper]{article}

\usepackage{cvpr}
\usepackage{times}
\usepackage{graphicx}
\usepackage{amsmath}
\usepackage{amssymb}

\usepackage{stfloats}
\usepackage{array}
\usepackage{multirow}
\usepackage{subcaption}
\usepackage[breaklinks=true,bookmarks=false]{hyperref}
\cvprfinalcopy
\providecommand{\e}[1]{\ensuremath{\times 10^{#1}}}
\def\cvprPaperID{****}

\begin{document}

\title{Recurrent Attention Models for Depth-Based Person Identification}

\author{Albert Haque, Alexandre Alahi, Li Fei-Fei\\
	Computer Science Department, Stanford University\\
	{\tt\small \{ahaque,alahi,feifeili\}@cs.stanford.edu}
}

\maketitle

\begin{abstract}
We present an attention-based model that reasons on human body shape and motion dynamics to identify individuals in the absence of RGB information, hence in the dark. Our approach leverages unique 4D spatio-temporal signatures to address the identification problem across days.
Formulated as a reinforcement learning task, our model is based on a combination of convolutional and recurrent neural networks with the goal of identifying small, discriminative regions indicative of human identity. We demonstrate that our model produces state-of-the-art results on several published datasets given only depth images.
We further study the robustness of our model towards viewpoint, appearance, and volumetric changes.
Finally, we share insights gleaned from interpretable 2D, 3D, and 4D visualizations of our model's spatio-temporal attention.
\end{abstract}

\section{Introduction}

A quick, partial view of a person is often sufficient for a human to recognize an individual. This remarkable ability has proven to be an elusive task for modern computer vision systems.
Nevertheless, it represents a valuable task for security authentication, human tracking, public safety, and role-based activity understanding \cite{kale2003gait, ioannidis2007gait, alahi14}.

Given an input image, person identification aims to assign identification labels to individuals present in the image.
Despite the best efforts from previous work \cite{zhao2014learning, zheng2015partial, li2013locally}, this problem remains largely unsolved.
Without accurate spatial or temporal constraints, visual features alone are often intrinsically weak for matching people across time due to intra-class differences. Additional variances due to illumination, viewpoint, and pose further exacerbate the problem.

\begin{figure}[t]
	\centering
	\includegraphics[width=1\linewidth]{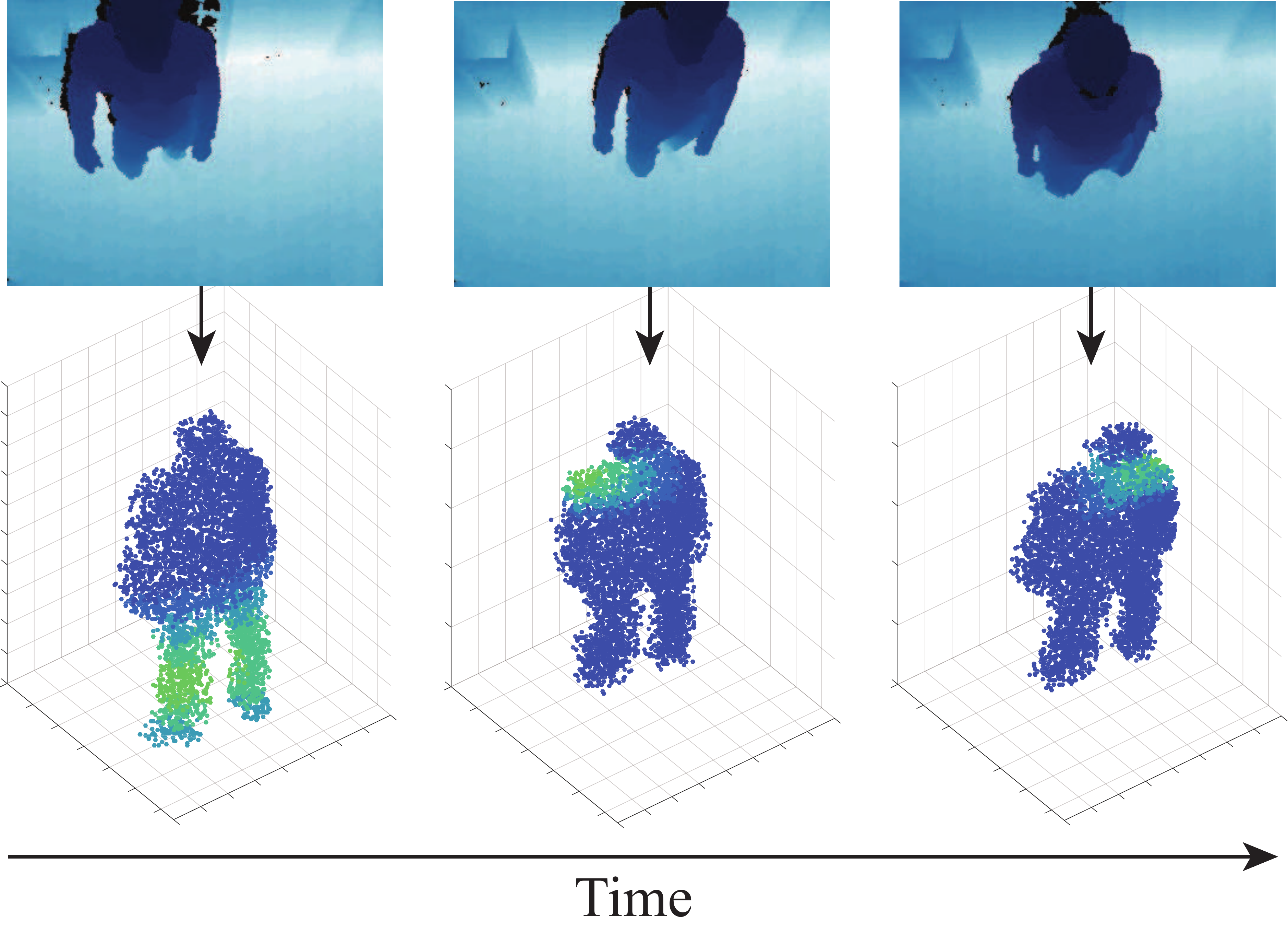}
	\caption{Gait has been shown to be unique to each person. We propose a 4D recurrent attention model to learn spatio-temporal signatures and identify people from depth images.}
	\label{fig:pull}
\end{figure}

Research findings from physiology and psychology have shown that gait is unique to each individual \cite{murray1964walking, murray1967gait, cutting1977recognizing}. Building on this observation, we aim to learn body shape and motion signatures unique to each person (see Figure \ref{fig:pull}). Inspired by the recent success of the depth modality \cite{andersson2014anthropometric, zhao20063d}, our goal is to output an identification label from a depth image or video.

The primary challenge towards this goal is designing a model that is not only rich enough to reason about motion and body shape but also robust to intra-class variability.
The second challenge is that person identification inherently comprises of a large number of classes with few training examples per class (in some cases a single training example).
Existing datasets \cite{munaro20143d, barbosa12reid, munaro2014feature} often collect front-facing views with constant appearances (\ie similar sets of clothing).
While this makes the identification problem more tractable, we are interested in relaxing these assumptions to solve a more general identification task which is applicable to a broader audience.

Our core insight is that we can leverage raw depth video despite the scarcity of training inputs, to address the aforementioned challenges by formulating the task as a reinforcement learning problem. Our approach involves pruning the high dimensional input space and focuses on small, discriminative regions while being free of visual and temporal assumptions. Concretely, our contributions are:

(i) We develop a recurrent attention model that identifies humans based on depth videos. Our model leverages a 4D input and is robust to appearance and volumetric changes.
By combining a sparsification technique with a reinforcement learning objective, our recurrent attention model attends to small spatio-temporal regions with high fidelity while avoiding areas with little information (see Section \ref{sec:model}).

(ii) We re-examine the person identification task and build a challenging dataset which taxes existing methods (see Section \ref{sec:perf}). We push the limits of our model by varying the viewing angle and testing on diverse training examples of people carrying objects (\eg coffee or laptops) or wearings hats and backpacks.

In Section \ref{sec:perf}, we show that our model achieves state-of-the-art results on several existing datasets. Furthermore, we take advantage of our recurrent attention model and create interpretable 2D, 3D, and 4D visualizations of hard attention \cite{xu2015show}. Our findings shed new insights on volumetric and motion-based differences between individuals.
To aid in future research, we make all code, data, and annotations publicly available upon publication.

\section{Related Work}

\textbf{RGB-Based Methods.} The primary challenge associated with identification is intra-class variance. These include changes in appearance due to illumination, point of view, pose, and occlusion. There have been many attempts to solve this problem by improving the feature representations \cite{gray2008viewpoint, wang2007shape, farenzena2010person, zhao2013unsupervised, zhao2014learning, kviatkovsky2013color, zheng2015partial} and by exploring new similarity metrics \cite{li2013locally, mignon2012pcca, prosser2010person}. Silhouette-based approaches ignore color altogether and use anthropometric or geodesic distances between body parts \cite{kale2003gait, mansur2014gait, skog2010gait}.

\textbf{Depth-Based Methods.}
Following suit from silhouette-based approaches, several depth-based studies have applied anthropometric and soft biometrics to the 3D human skeleton \cite{mogelmose2013multimodal, albiol2012different, munsell2012person, andersson2014anthropometric, dubois2014gait}. Harnessing the full power of depth cameras, several papers investigated 3D point clouds for person identification \cite{zhao20063d, ioannidis2007gait}. Although these approaches are successful, they rely on hand-crafted features (\eg arm length, torso width) or low-level RGB features (\eg SURF \cite{bay2006surf}, SIFT \cite{lowe1999object}).

\textbf{Spatio-Temporal Representations}.
Methods described thus far have largely ignored spatio-temporal information. Originally proposed in \cite{han2006individual}, the gait energy image and ifigts variants \cite{chunli2010behavior, bashir2010gait, huang2010gait, veeraraghavan2005matching}, embed temporal information onto a two-dimensional image by averaging the silhouette across all frames of a video. Test time predictions are obtained from a $k$-nearest neighbor lookup.

More recently, the gait energy image has been extended into 3D by using depth sensors \cite{hofmann20122, sivapalan2011gait}. Spatial volumes and higher-dimensional tensors have been proposed for activity and action recognition \cite{oreifej2013hon4d, wang2012robust, zanfir2013moving, blank2005actions, jhuang2013towards, laptev2007local}, medical image analysis \cite{shin2013stacked}, robotics \cite{maturana_icra_2014, maturana_iros_2015}, and human motion analysis \cite{laptev2005space} but have not been thoroughly explored in the person identification domain.

\textbf{Deep Learning for Identification.}
A small number of studies have explored the applicability of deep neural networks to person identification. In \cite{yi2014deep}, Yi et al. proposed a siamese convolutional neural network for similarity metric learning. In \cite{li2014deepreid}, Li et al. proposed a similar approach by using filter pairs to model photometric and geometric transforms. Following these works, Ding et al. \cite{ding2015deep} formulated the input as a triplet containing both correct and incorrect reference images. In \cite{ahmed2015improved}, Ahmed et al. introduced cross-input neighborhood differences.

Our work has several key differences with the aforementioned works: First, we focus on the depth modality and do not use any RGB information. Second, the methods above \cite{yi2014deep, li2014deepreid, ding2015deep, ahmed2015improved} ingest several images as input and compute similarity between these inputs. They formulate the identification problem as an image-similarity task using images captured from non-overlapping camera views. Our model uses a single image\footnote{The input to our model is one image for frame-wise identification or one sequence for video-level (i.e. temporal or voting) identification.} as input and does not rely on metric learning.

\textbf{Attention Models}. Interpretability of deep learning models is becoming increasingly important within the machine learning and computer vision communities.

By measuring the sensitivity of output variables to variances in the input, attention models applied to image classification \cite{zeiler2014visualizing, gregor2015draw, xiao2014application}, image captioning \cite{fang2014captions, xu2015show, chen2015mind}, object detection \cite{caicedo2015semantic}, and tracking \cite{denil2012learning} have demystified many aspects of convolutional and recurrent networks. These methods exploit the spatial structure of the input to understand intermediate network representations. Sequential data, on the other hand, requires temporal attention models to understand the order dependence of the input data. Recent papers in speech recognition \cite{graves2006connectionist}, video captioning \cite{yao2015describing}, and natural language processing \cite{karpathy2015visualizing, luong2015effective, chan2015listen} explore the concept of attention in the temporal domain.

Many deep learning models impose constraints on the input. Due to the high dimensionality of images (\ie high pixel count), preprocessing often includes resizing and/or cropping the original input image \cite{krizhevsky2012imagenet}. Videos are often truncated to a fixed length for training. Due to computational limitations, this loss of information is necessary to constrain runtimes. In the next section, we describe our model and how we balance this trade-off by employing visual ``glimpses" \cite{mnih2014recurrent} which process small 4D regions with high fidelity and grow to larger regions with lower detail.

\section{Our Model}\label{sec:model}

\begin{figure*}[t!]
	\centering
	\vspace{-5mm}
	\includegraphics[width=1.0\linewidth]{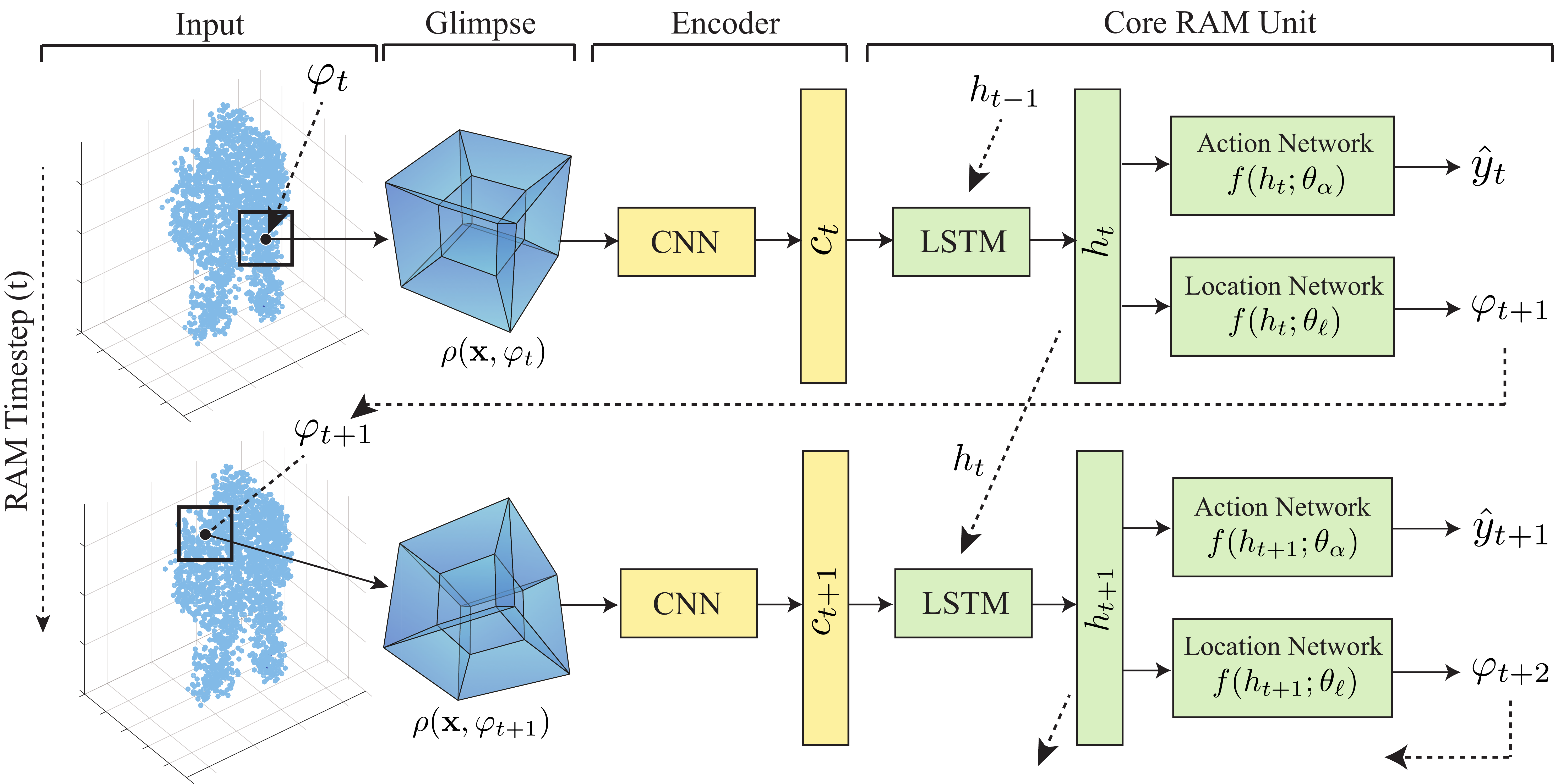}
	\vspace{-3mm}
	\caption{Our full model. Dashed arrows indicate information exchange across time steps. Solid arrows indicate information exchange within a time step. Two time steps are shown with a series of events occurring from left to right.  Note: RAM timestep $t$ refers to the ``iteration" of our model and does not refer to the input video timestamp $\tau$. All other variables are defined in Section \ref{sec:ram}.
	}\label{fig:model}
\end{figure*}

The goal of our model is to identify humans from depth images or video. Our model (Figure \ref{fig:model}) computes hard attention regions \cite{xu2015show} which are used to predict an identification label. In this section, we describe our 4D input representation followed by a discussion of our attention model.

\subsection{Input Representation}
Projections from higher dimensional spaces onto lower spaces result in information loss. This serves as our motivation for using 4D data: we want to preserve as much information as possible and let our model decide the relevant regions. Four-dimensional data consists of a 3D point cloud (\eg, $x,y$, and $z$ cooridnate) and time $\tau$. For simplicitly, Figure \ref{fig:model} shows the input as 3D point clouds which are constructed from depth images.

Each training example ($\mathbf{x}, \mathbf{y})$ consists of a variable sized 4D tensor $\mathbf{x}$ and corresponding label $\mathbf{y}$. The tensor is variable due to variable video lengths. Let $f$ denote the number of frames in video $i$ and let $x,y$ and $z$ denote the width, height, and depth dimensions of our tensor.\footnote{We use a tensor of size $250 \times 100 \times 200$.} 
\begin{equation}
\mathbf{x} \in \mathbb{R}^{f \times x \times y \times z} \textrm{\quad and \quad} \mathbf{y} \in [1,...,C]
\end{equation}
where $C$ is the number of classes. For an average video containing 500 frames, flattening $\mathbf{x}$ leads to a feature vector of 2.5\e{9} elements. For comparison, a $227 \times 227$ RGB image (typical for a convolutional network), results in 1.2\e{6} elements. This means that our model must operate on an input space three orders of magnitude larger than common convolutional networks. Consequently, our model must be designed to intelligently navigate this high dimensional space.

\subsection{Recurrent Attention Model}\label{sec:ram}

Given this high-dimensional depth representation, we want our model to focus on smaller, discriminative regions in the input space. Minh et al. \cite{mnih2014recurrent} recently proposed the recurrent attention model (RAM) for image classification and reinforcement learning problems. While they show promising results, they enjoyed several advantages. First, training data is plentiful. Image classification has been well-studied and several large benchmarks exist. Dynamic environments such as a control-based video game can generate data on-the-fly as the game is played. Second, the input dimensionality of these problems is relatively small: MNIST is $28 \times 28$ while the control game is $24 \times 24$  \cite{mnih2014recurrent}.

Person identification, on the other hand, does not enjoy these advantages. Instead, we are tasked with limited, high-dimensional training data. Figure \ref{fig:model} shows an overview of our proposed model. It consists of a glimpse layer which down-samples the input, an encoding stage which acts as an additional dimensionality reduction tool, and a core RAM network responsible for spatio-temporal learning.

\textbf{Glimpse Layer.} The goal of the glimpse layer is two-fold: (i) it must avoid (or greatly limit) information loss and (ii) it must refrain from processing large inputs.
At a given time step $t$, our model does not have full access to the input $\mathbf{x}$ but instead extracts a partial observation or ``glimpse" denoted by $\rho(\mathbf{x}, \varphi_t)$.
A glimpse encodes the region around $\varphi_t$ with high resolution but uses a progressively lower resolution for points further from $\varphi_t$. Adopting a multi-scale strategy has been shown to be an effective de-noising technique \cite{zontak2013separating}.
Additionally, this results in a tensor with much lower dimensionality than the original input $\mathbf{x}$.
By focusing on specific regions, we can reduce the required computation by our model, reducing the loss of spatio-temporal detail, and reduce the effect of noise.

As shown in Figure \ref{fig:model}, a glimpse is comprised of $G$ hypercube patches. The first patch has a side length of $g_s$ and maintains full resolution of the input centered at $\varphi$.
The second patch has a side length of $2g_s$ and is sampled at $1/2$ resolution. Patches grow in size with progressively lower resolution. Specifically, the $k^{th}$ patch has a side length of $kg_s$ and is sampled at $1/k$ of the original input resolution. The final glimpse is a concatenation of these hypercube patches.

\textbf{Encoder}. The glimpse still contains a large number of features (on the order of 1\e{6}). We must further compress the glimipse before it becomes a feasible solution for our data-limited person identification task.
To accomplish this, we use an encoding layer to further reduce the feature space. In our model, this is done with a 4D convolutional autoencoder  \cite{masci2011stacked, ji20133d}.
The encoder layer is trained offline and separately from the RAM. During RAM training and test time, encoded features are denoted as $c_t$.

\textbf{Core RAM Unit}. As mentioned previously, the number of features associated with a 4D input is on the order of 1\e9. Conventional deep learning methods cannot feasibly explore and learn from the full input space. Motivated by this, we use a recurrent attention model. Our goals of the RAM are two-fold: First, model interpretability is an overarching theme of this work. Given image-based input, an attention-based model allows us to visually understand human shape and body dynamics. Second, a RAM provides us with computational advantages by pruning the input space by focusing on rich, discriminative regions.

As shown in Figure \ref{fig:model} our model is a recurrent network: it consists of a long short-term memory (LSTM) unit \cite{hochreiter1997long} and two sub-networks. Parameterized by $\theta_r$, our LSTM receives encoded features $c_t$ and the previous hidden layer $h_{t-1}$ at each time step $t$ and outputs a hidden state $h_t$.

\textbf{Sub-Networks.} Before the next iteration of our RAM, our model must take two actions: (i) it decides the next glimpse location and (ii) it outputs a predicted identification label for the current time step. We compute these by using two sub-networks: the location and action network, respectively.

The location network stochastically selects the next glimpse location using the distribution parametrized by $f(h_t; \theta_\ell)$ (where $\theta_\ell$ refers to the location network's parameters). Similar to \cite{mnih2014recurrent}, the location network outputs the mean of the location policy (defined by a 4-component Gaussian) at time $t$ and is defined by: $ f(h_t; \theta_\ell) = tanh(Linear(h_t)) $ where $Linear(\bullet)$ is a linear transformation.

The action network (parameterized by $\theta_\alpha$), outputs a predicted class label $\hat{y}$ given the current LSTM hidden state, $h_t$. Parameterized by $f(h_t; \theta_\alpha)$, the action network consists of a linear and softmax layer defined by $f(h_t; \theta_\alpha) =\exp(Linear(h_t)) / Z$ where $Z$ is a normalizing factor. The predicted class label $\hat{y}_t$ is then selected from the softmax output.

\subsection{Training and Optimization}

\textbf{Formulation.} Depth video is inherently a large feature space. To avoid exploring the entire input space, we pose the training task as a reinforcement learning problem. After our model decides the label $\hat{y}$ and next glimpse location $\varphi$, our model receives a reward $R$ where $R=1$ if $\hat{y}_t=\mathbf{y}$ at time $T$, where $T$ is a threshold for the maximum number of time steps; otherwise $R=0$. Let $\Theta = \{\theta_r, \theta_\ell, \theta_\alpha\}$ denote all parameters of the RAM.

Let $s_{1:t} = \mathbf{x},\varphi_1,\hat{y}_1, ..., \mathbf{x},\varphi_t,\hat{y}_t$ denote the historical sequence of all input-action pairs (\ie input tensor, predicted label, and next glimpse). We call this a \textit{glimpse path}. A glimpse path shows where our model ``looks at" over time\footnote{For 4D input, time refers to the iteration of the RAM and not the input video's frame order.}. Our model must learn a stochastic policy $\pi(\varphi_t, \hat{y}_t | s_{1:t}; \Theta)$ which maps the glimpse path $s_{1:t}$ to a distribution over actions for the current time step. The policy $\pi$ is defined by our core RAM unit and the history $s_t$ is embedded in the LSTM's hidden state $h_t$.

\textbf{Optimization.} The policy of our model induces a distribution over possible glimpse paths. Our goal is to maximize the reward function over $s_{1:N}$:
\begin{equation}\label{eq:reward_function}
J(\Theta) =  \mathbb{E}_{p(s_{1:t};\Theta)} \left[ R \right]
\end{equation}
where $p(s_{1:T};\Theta)$ depends on the policy $\pi$.
However, computing the expectation introduces unknown environment parameters which makes the problem intractable. Formulating the task as a partially-observable Markov decision process allows us to compute a sample approximation to the gradient, known as the REINFORCE rule \cite{williams1992simple}:
\begin{align}
\nabla_\Theta J(\Theta) &= \sum\limits_{t=1}^{T} \mathbb{E}_{p(s_{1:T};\Theta)} \Big( \nabla_\Theta \log \pi(\mathbf{y}| s_{1:t};\Theta) R \Big) \\
&\approx \frac{1}{M} \sum\limits_{i=1}^{M} \sum\limits_{t=1}^{T} \nabla_\Theta \log \pi(\mathbf{y} | s^{(i)}_{1:t}; \Theta) R^{(i)} \label{eq:reinforce}
\end{align}
where $s^{(i)}_{1:t}$ denotes the glimpse path, $R^{(i)}$ denotes the reward, and $\mathbf{y}^{(i)}$ denotes the correct label for the $i^{th}$ training example. Additionally, $\nabla_\theta \log \pi(u_t^{(i)} | s^{(i)}_{1:t}; \theta) R^{(i)}$ is the gradient of the LSTM. Consistent with \cite{mnih2014recurrent}, we train the action network with the cross entropy loss function and train the location network with REINFORCE. This formulation allows our model to focus on salient 3D regions in both space and time.

\textbf{Advantages.} 
A major benefit of this formulation is that limited training data is no longer an issue. Our model is trained on glimpses (\ie subsets of the input) and not the entire video sequence. Therefore, the effective number of training examples made available to our model is on the order of 1\e6 to 1\e9 per video (\ie number of possible glimpses). Despite having a single video as input, our model almost never sees the same training example twice. Our model is still limited by the number of training data but our formulation makes it less of a concern.

\section{Experiments}\label{sec:perf}

First, we describe our datasets and evaluation metrics. This is followed by a discussion of experimental, hyperparameter, and design selections. We then present results for the single-shot (single image) and multi-shot (multi-frame) person identification task. We then show 2D, 3D, and 4D visualizations followed by concluding remarks on our model's limitations.

\subsection{Datasets}\label{sec:datasets}

\begin{table}[t]
	\centering
	\vspace{-4mm}
	\footnotesize
	\begin{tabular}{l|cccc}
		& \textbf{BIWI}  & \textbf{IAS-A/B}  & \textbf{PAVIS}  & \textbf{DPI-T} \\ \hline
		\# Unique Subjects & 50 (28)     & 11 (11)               & \textbf{79 (79)}        & 12 (12) \\
		\# Total Videos   & 50 (56)       & 11 (11)            & 79 (79)        & \textbf{300 (355)}\\
		\# Appearances/class  & 1 (2) & 1 (2) & 1 (2) & \textbf{5 (5)} \\
		\# 2D inputs/class & 479 (551) & \textbf{701 (739)} &  5 (5) & 336 (398) \\
		\# 3D inputs/class & 479 (551) & \textbf{701 (739)} & 5 (5) & 336 (398) \\
		\# 4D inputs/class  &1 (2) & 1 (1) & 1 (1) & \textbf{ 25 (30)} \\ \hline
	\end{tabular}
	\caption{Comparison of datasets. DPI-T is our newly collected dataset. We list the number of subjects, images, and videos for both the training and test sets. The test set is shown in parenthesis.
		Appearance is defined by a person wearing unique clothing or distinct visual appearance.}
	\label{table:datasets}
\end{table}
Our goal is to identify humans based on their 3D shape and body dynamics captured by a depth camera. The majority of human-based RGB-D datasets are catered to human activity analysis and action recognition \cite{cao2010cross, chen2015mhad, h36m_pami}. Since they generally consist of many gestures performed by few subjects, these datasets are not suited for the identification problem. We hence use existing depth-based identification datasets and collected a new one to further test our model.

We evaluate our model on several existing depth-based identification datasets: BIWI \cite{munaro20143d}, IIT PAVIS \cite{barbosa12reid}, and IAS-Lab \cite{munaro2014feature}. These datasets contain 50, 79, and 11 humans, respectively. For BIWI, we use the full training set and the \textit{Walking} test set. For PAVIS, we use \textit{Walking1} and \textit{Walking2} as the training and test set, respectively. For IAS-Lab, we use the full training set and both test splits.

Existing datasets impose constraints to simplify the identification problem (\eg, few sets of clothing per person, front-facing views, or slow walking speed). We collected a new dataset: Depth-Based Person Identification from Top (DPI-T), which is different from previous datasets.

\begin{figure}[t]
	\centering
	\vspace{-1mm}
	\includegraphics[width=1.0\linewidth]{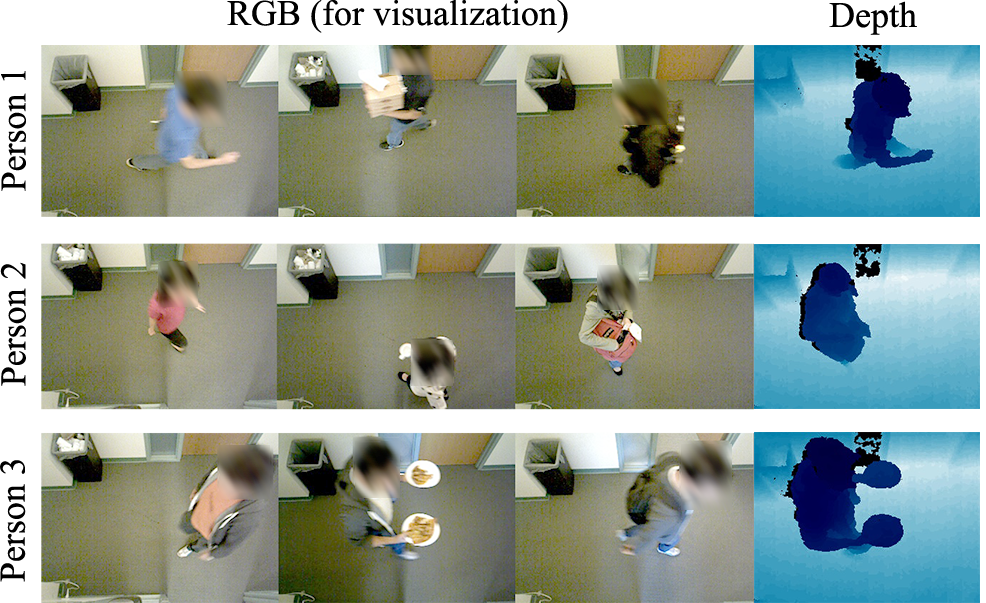}
	\vspace{-2mm}
	\footnotesize
	\caption{Sample images from our Depth-Based Person Identification from Top (DPI-T) dataset. Each row denotes a different person. The three left columns show RGB images for convenience. Our model only uses depth images, as depicted in the right column.
	}\label{fig:ours}
\end{figure}

\textbf{We provide more observations per individual.} On average, individuals appear in a total of 25 videos across several days. This naturally results in individuals wearing different sets of clothing -- 5 different sets of clothing on average. Figure \ref{fig:ours} shows three individuals from our dataset wearing different sets of clothing. Additionally, people in our dataset walk at variable speeds depending on the time of day or week.

\begin{table*}[t]
	\centering
	\footnotesize
	\vspace{-6mm}
	\begin{tabular}{lll|ccccc|ccccc}
		\hline
		& & & \multicolumn{5}{c|}{\textbf{Top-1 Recognition Rate (\%)}} & \multicolumn{5}{c}{\textbf{Normalized Area Under the Curve (nAUC)}} \\ \hline
		\textbf{\#} & \textbf{Modality} &  \textbf{Methods} & \textbf{BIWI} & \textbf{IAS-A} & \textbf{IAS-B} & \textbf{PAVIS} & \textbf{DPI-T} & \textbf{BIWI} & \textbf{IAS-A} & \textbf{IAS-B} & \textbf{PAVIS} & \textbf{DPI-T} \\ \hline
		1 & Depth & Random  & 2.0 & 9.1 & 8.1 & 1.3 & 8.3 & 51.0 & 54.5 & 54.5 & 50.6 & 54.2 \\
		2 & Depth & Human Performance & 6.7 & 21.2 & 15.1 & 1.7  & 19.2 & --- & --- & --- & --- & --- \\  \hline
		3 & Depth &  Skeleton (NN) \cite{barbosa12reid} & --- & --- & --- & 15.0 & --- & --- & --- & --- & 91.8 & --- \\
		4 & Depth &  Skeleton (NN) \cite{munaro20143d} & 21.1 & 22.5 & 55.5 & 28.6 & --- & 81.7 & 72.8 & 86.3 & 89.9 & --- \\
		5 & Depth &  Skeleton (SVM) \cite{munaro2014one} & 13.8 & --- & --- & 35.7 & --- & 86.6 & --- & --- & 92.8 & --- \\
		6 & Depth & 3D CNN           & 27.7 & 44.2 & 56.2 & 27.5 & 23.7    & 88.2 & 86.1 & 86.0 & 89.2 & 75.6 \\
		7 & Depth &  2D RAM         & 24.7 & 46.9 & 61.0 & 30.5 & 33.8     & 87.4 & 87.7 & 86.8 & 90.1 & 82.5 \\
		8 & Depth &  3D RAM         & \textbf{30.1} & \textbf{48.3} & \textbf{63.7} & \textbf{41.3} & \textbf{47.5 }   & \textbf{88.7} & \textbf{88.5} & \textbf{87.7} & \textbf{93.7} & \textbf{88.3} \\  \hline
		9* & RGB & Face Detection \cite{munaro20143d} & 36.7* & --- & --- & --- & --- & 87.6* & --- & --- & --- & --- \\
		10* & RGB  & PTZ Max-Var \cite{salvagnini2013person} & --- & --- & --- & 73.1* & --- & --- & --- & --- & 98.7* & --- \\
		11* & RGB-D & Face+Skeleton \cite{munaro2014one} & 43.9* & --- & --- & --- & --- & 90.2* & --- & --- & --- & --- \\
		12* & RGB-D  & PCM+Skeleton \cite{munaro20143d} & 27.4* & 25.6* & 63.3* & --- & --- & 87.4* & 75.5* & 86.3* & --- & --- \\ \hline
	\end{tabular}
	
	\caption{Single-shot identification performance. Methods shown above use only spatial information. A summary of each method can be found in Section \ref{sec:baselines}. Both metrics were computed on the test set. Larger values are better. Dashes indicate that no published information is available. (*) Although not a fair comparison, for sake of completeness, we list RGB and RGB-D methods.}
	\label{table:results_single}
\end{table*}

\begin{figure*}[t!]
	\centering
	\vspace{-3mm}
	\begin{subfigure}[t]{0.325\linewidth}
		\centering
		\includegraphics[width=\linewidth]{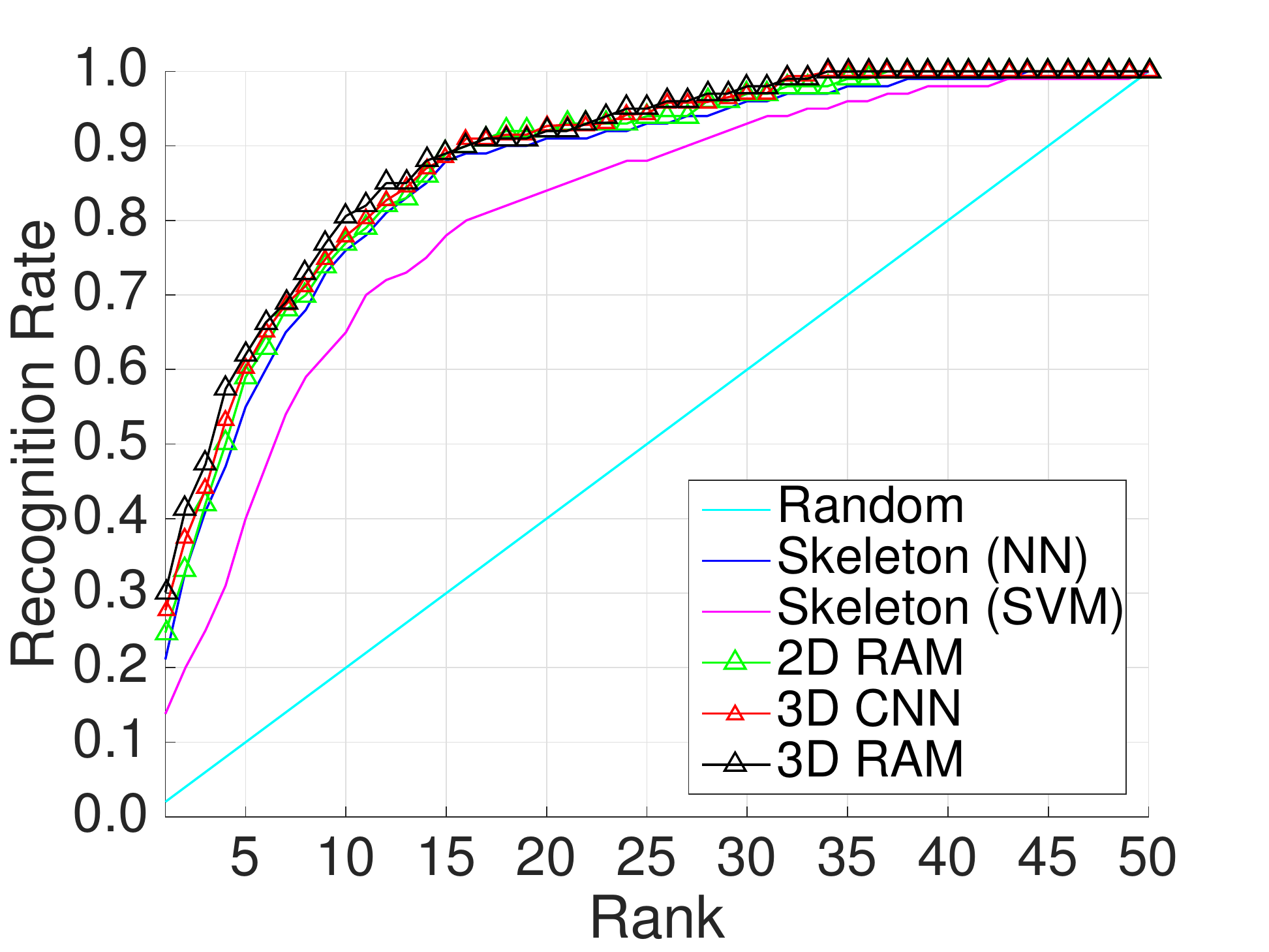}
		\caption{BIWI (Single-Shot)}
	\end{subfigure}%
	~
	\begin{subfigure}[t]{0.325\linewidth}
		\centering
		\includegraphics[width=\linewidth]{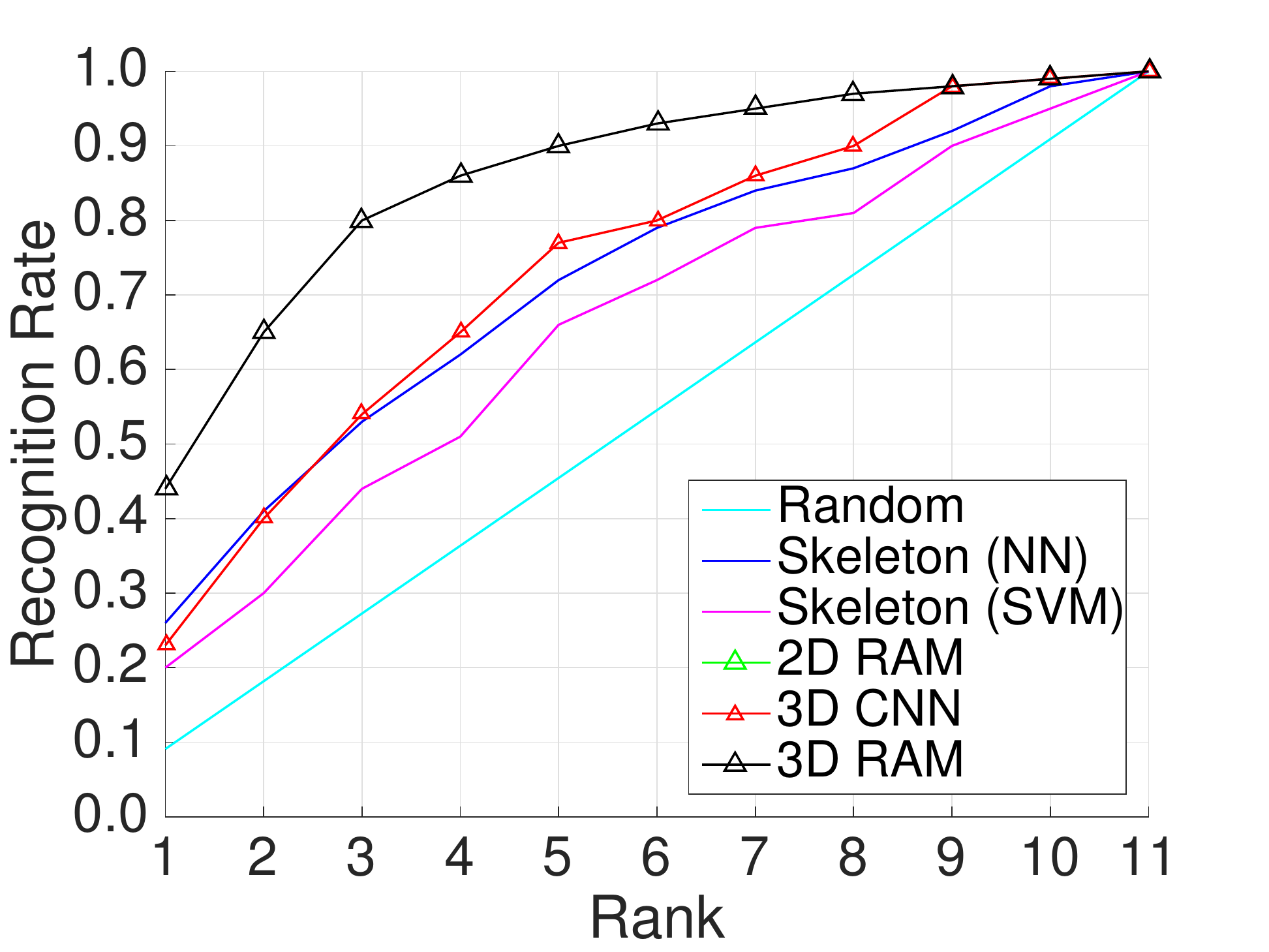}
		\caption{IAS-Lab-A (Single-Shot)}
	\end{subfigure}%
	~
	\begin{subfigure}[t]{0.325\linewidth}
		\centering
		\includegraphics[width=\linewidth]{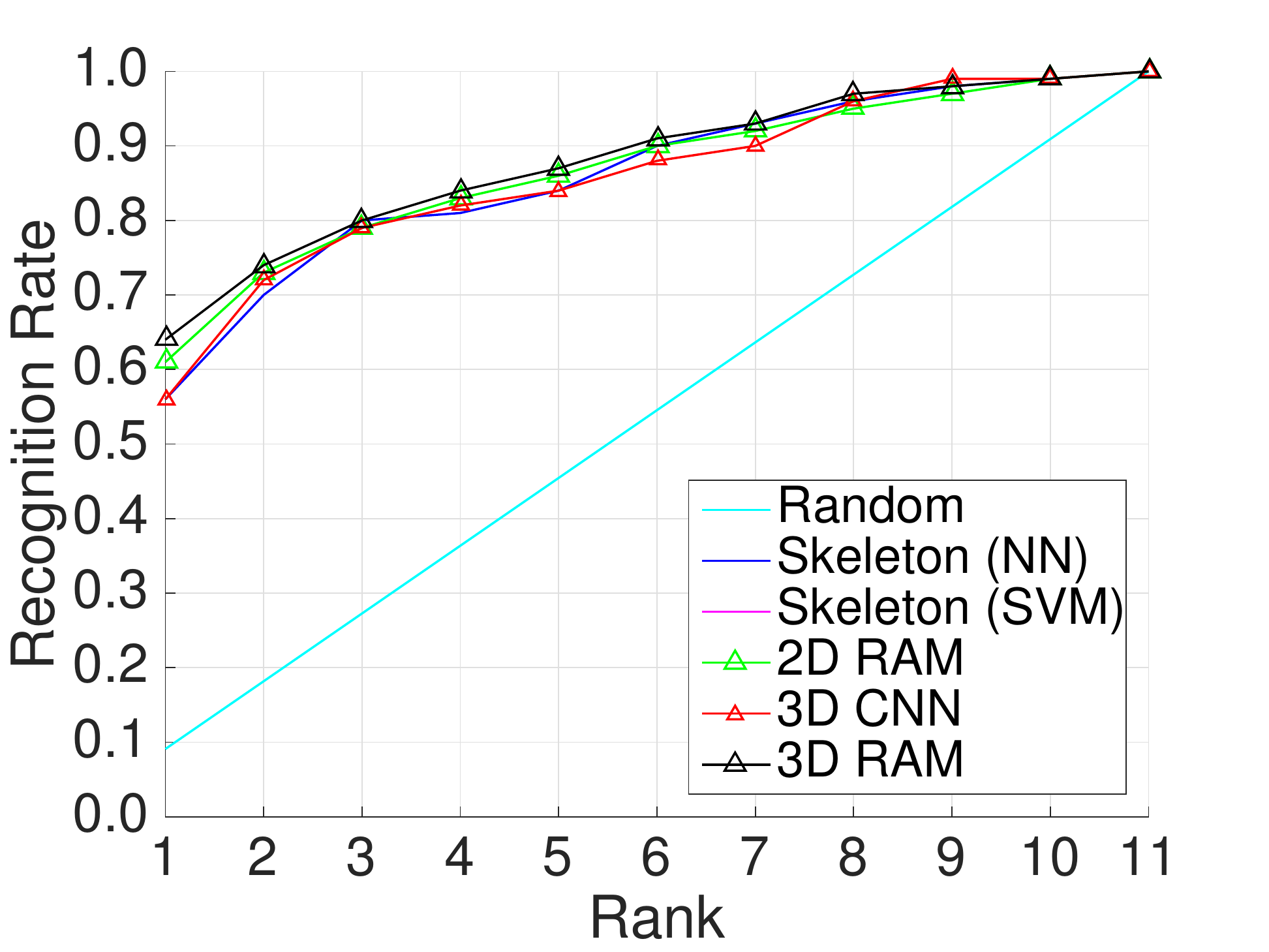}
		\caption{IAS-Lab-B (Single-Shot)}
	\end{subfigure}%
	
	\caption{(a-c) Cumulative matching curves for test set performance on various datasets and models. Dataset details can be found in Section \ref{sec:datasets}. Model details can be found in Section \ref{sec:baselines}. The $y$ axis denotes recognition rate. For the $x$ axis, rank-$k$ is the recognition rate if the ground truth label is within the model's top-$k$ predictions.}
	\label{fig:cmc}
\end{figure*}

\textbf{Challenging top-view angles.} In real-world applications such as smart spaces and public environments (\eg, hospitals, retail stores), cameras are often attached to the ceiling pointed down, as opposed to clean, frontal or side view images available in existing datasets. This introduces self-occlusion challenges and often leads to undetected faces and incomplete 3D point cloud reconstructions.

\textbf{People are holding objects.} Existing datasets collect data from the simple case of walking in a controlled environment. In our dataset, people are ``in the wild," often holding objects such as coffee, laptops, or food. Additionally, since our dataset is collected across a long period of time, people often wear hats, bags, or carry umbrellas (see Figure \ref{fig:ours}). A table showing the characteristics of existing datasets and our new dataset is shown in Table \ref{table:datasets}.

\subsection{Evaluation Metrics}
Person identification can be solved in a ``single-shot" manner using one image to produce a label or a ``multi-shot" method which leverages multiple frames, temporal features, or multi-frame voting schemes. Below, we provide evaluation results for both single-shot and multi-shot approaches.

Specific metrics include the top-1 recognition rate, cumulative matching curve (CMC), normalized area under the curve (nAUC) metrics. Top-$k$ recognition rate indicates the fraction of test examples that contained the ground truth label within the top-$k$ predictions. Generalizing the top-$k$ metric to higher ranks (up to the number of people in the dataset), produces the cumulative matching curve. Integrating the area under the CMC curve and normalizing for the number of ranks produces the nAUC.

\subsection{Experimental Settings}
Tensors were fixed to a size of $250 \times 100 \times 200$ and converted to integer indices corresponding to the $x,y$, and $z$ real world coordinates. The $x$ and $y$ units represent real world centimeters while the $z$ units represents 10 millimeters. 
Glimpse locations are encoded as $\varphi = (x,y,z,\tau)$ where $x,y,z$ are real values while $\tau$ is integer valued.
The first glimpse patch has a side length of $8$ tensor units and we use 5 glimpse patches. 
For 3D and 4D inputs, we augment the data by applying Gaussian noise, with mean of 0 cm and 5 cm variance, to each point in the point cloud. Images and tensors are shifted between 0 and $\pm 5$ cm in all directions about the origin and randomly scaled between $0.8\times$ and $1.2\times$.
We train our model from scratch using stochastic gradient descent with mini-batches of size 20, a learning rate of 1\e{-4}, momentum of 0.9, and weight decay of 5\e{-4}. The CNN was pretrained on augmented training examples before RAM training. All learning layers employ dropout \cite{srivastava2014dropout} with 0.5 probability. 

\begin{table*}
	\centering
	\vspace{-6mm}
	\footnotesize
	\begin{tabular}{lll|ccccc|ccccc}
		\hline
		& & & \multicolumn{5}{c|}{\textbf{Top-1 Recognition Rate (\%)}} & \multicolumn{5}{c}{\textbf{Normalized Area Under the Curve (nAUC)}} \\ \hline
		\textbf{\#} & \textbf{Modality} &  \textbf{Methods} & \textbf{BIWI} & \textbf{IAS-A} & \textbf{IAS-B} & \textbf{PAVIS} & \textbf{DPI-T} & \textbf{BIWI} & \textbf{IAS-A} & \textbf{IAS-B} & \textbf{PAVIS} & \textbf{DPI-T} \\ \hline
		1 & Depth & Random  & 2.0 & 9.1 & 8.1 & 1.3 & 8.3 & 51.0 & 54.5 & 54.5 & 50.6 & 54.2 \\
		2 & Depth & Human Performance & 6.7 & 21.2 & 15.1 & 1.7  & 19.2 & --- & --- & --- & --- & --- \\  \hline
		3 & Depth &  Energy Image \cite{chunli2010behavior} & 21.4 & 25.6 & 15.9 & 29.1 & 18.5 & 73.2 & 72.1 & 66.0 & 81.2 & 75.8 \\
		4 & Depth &  Energy Volume \cite{sivapalan2011gait} & 25.7 & 20.4 & 13.7 & 18.9 & 14.2 & 83.2 & 66.2 & 64.8 & 68.3 & 65.5 \\
		5 & Depth &  Skeleton (NN) \cite{munaro2014one} & 39.3 & --- & --- & --- & --- & --- & --- & --- & --- & --- \\
		6 & Depth &  Skeleton (SVM) \cite{munaro2014one} & 17.9 & --- & --- & --- & --- & --- & --- & --- & --- & --- \\
		7 & Depth & Skeleton (LSTM) & 15.8 & 20.0 & 19.1 & 14.5 & --- & 65.8 & 65.9 & 68.4 & 64.0 & --- \\
		8 & Depth & 3D CNN+Avg Pooling \cite{boureau2010theoretical} & 27.8 & 33.4 & 39.1 & 27.5 & 28.4 & 84.0 & 81.4 & 82.8 & 80.6 & 82.5 \\ 
		9 & Depth &  3D LSTM        & 27.0 & 31.0 & 33.8 & 20.3 & 23.9  & 83.3 & 77.6 & 78.0 & 77.1 & 77.9 \\
		10 & Depth &  4D RAM         & \textbf{45.3} & \textbf{53.5} & \textbf{64.4} & \textbf{43.0} & \textbf{55.6}  & \textbf{91.2} &\textbf{ 91.4} & \textbf{89.0} & \textbf{93.4} & \textbf{91.6} \\ \hline
		11* & RGB & Face Detection \cite{munaro20143d} & 57.1* & --- & --- & --- & --- & --- & --- & --- & --- & --- \\
		12* & RGB-D  & Face+Skeleton \cite{munaro2014one} & 67.9* & --- & --- & --- & --- & --- & --- & --- & --- & --- \\
		13* & RGB-D & MCL+Skeleton \cite{pala2015multi} & --- & --- & --- & 89.0* & --- & --- & --- & --- & 98.9* & --- \\
		14* & RGB-D & PCM+Skeleton \cite{munaro20143d} & 42.9* & 27.3* & 81.8* & --- & --- & --- & --- & --- & --- & --- \\ \hline
	\end{tabular}
	\caption{Multi-shot identification performance. Methods shown above use multiple test images or use temporal information. A summary of each method can be found in Section \ref{sec:baselines}. Both metrics were computed on the test set. Larger values are better. Dashes indicate that no published information is available. (*) Although not a fair comparison, for sake of completeness, we list RGB and RGB-D methods.}
	\label{table:results_multi}
\end{table*}

\begin{figure*}[t]
	\centering
	\vspace{-2mm}
	\includegraphics[width=1\linewidth]{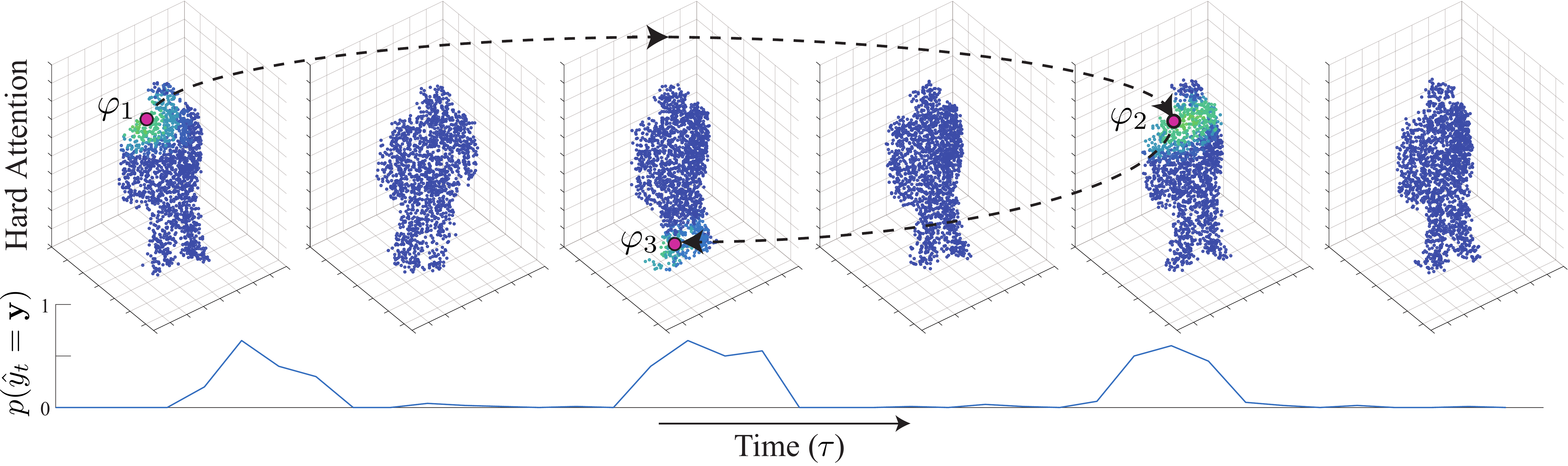}
	\vspace{-6mm}
	\caption{Hard 4D attention regions. Bright colored regions indicate areas closer to the glimpse center. Each point cloud is shown above in three dimensions ($x,y,z$) while the point clouds are arranged in video-order starting from the left. Arrows indicate jumps in time $\tau$. Although sparse point clouds are shown above, our model operates on the raw, dense point clouds.
	}\label{fig:attention_4d}
\end{figure*}

\subsection{Baselines}\label{sec:baselines}

\textbf{Single-Shot Identification.} We compare our recurrent attention model to several depth-based methods. Table \ref{table:results_single} shows various methods and results for the single-shot identification task: (1) We computed performance using a uniformly random guessing strategy. (2) Four humans manually performed the identification task. Each human was shown a single test input and was given full access to the training data. (3-5) Distances between skeleton joints are used as hand-crafted features \cite{barbosa12reid, munaro20143d, munaro2014one}. (6) A three-dimensional CNN operates on 3D point clouds. (7) A two-dimensional RAM operates on depth images. (8) A three-dimensional RAM operates on 3D point clouds. Although the focus of our paper is depth-based person identification, for completeness, we include related RGB and RGB-D methods to provide a more holistic view of the field. (9) A face descriptor is used \cite{munaro20143d}. (10) A point-tilt-zoom camera selectively zooms in on different parts of the image \cite{salvagnini2013person}. (11) A facial descriptor is concatenated with distances between skeleton joints \cite{munaro2014one}. (12) Similarity scores are computed on 3D point clouds and distances between skeleton joints \cite{munaro20143d}.

\textbf{Multi-Shot Identification.} Table \ref{table:results_multi} list several multi-shot methods. (1-2) We use random and human performance as baselines. (3-4) We evaluate the gait energy image \cite{han2006individual} and volume \cite{sivapalan2011gait}. (5-6) These methods use hand-crafted skeleton features with an inter-frame voting system. (7) Pairwise skeleton joint distances (same as 5-6) are fed into a LSTM. (8) A 3D CNN with average pooling \cite{boureau2010theoretical} over time \cite{yue2015beyond} . (9) A 3D LSTM operates on 3D point clouds. (10) Our final RAM model. (11-12) Face descriptors are used with a voting system. (13) A Multiple Component Dissimilarity (MCD) metric is computed on a pair of images. (14) RGB-D point cloud matching plus hand-crafted features are used for identification.

\subsection{Single-Shot Identification Performance}

\textbf{Learned encoding improves performance.} To better understand the source of our performance, we reduced the input dimensionality of our RAM and evaluated a 2D and 3D variant. The 2D and 3D models were evaluated on the single-shot task. As the dimensionality of the input increases from 2D to 3D, the performance of our RAM monotonically increases (see Figure \ref{fig:cmc}). Contrast this with gait energy in Table \ref{table:results_single}. Gait energy undergoes a similar transformation from 2D to 3D (\ie image to volume), but exhibits lower performance in the higher-dimensional case. This indicates that our learned encoder is able to preserve pertinent information from higher dimensional inputs whereas the gait energy volume fails without such encoding.

\textbf{RAM outperforms deep learning baselines.} As further validation of our model's performance, we evaluated a 3D convolutional neural network \cite{ji20133d}.
The input to both the 3D CNN and 3D RAM are 3D point clouds. As shown in Table \ref{table:results_single}, our 3D RAM model outperforms the 3D CNN. This confirms our hypothesis that our RAM is able to leverage glimpses to artificially increase the number of training examples and improve performance. The 3D CNN does not perform such data augmentation and instead operates on the entire point cloud.

\subsection{Multi-Shot Identification Performance}
Our final model (4D RAM) outperforms the human baseline and existing depth-based approaches. Both Munaro \etal \cite{munaro20143d} and Barbosa \etal \cite{barbosa12reid} used distances between skeleton joints as features. We list the performance of these hand-crafted features in Table \ref{table:results_multi}. Results show that these features are unable to infer the complex latent variables. Our 4D RAM model also outperforms an RGB-D method (13) in Table \ref{table:results_multi}. Proposed in \cite{munaro20143d}, method (13) computes a standardized 3D point cloud representation with the above skeleton-distance features. Although (13) leverages RGB information, it analyzes the entire point cloud which may include extraneous noise. Our model avoids noisy areas by selecting glimpses which contain useful information.

\subsection{Hard Attention Regions}

There is one key difference between our 3D and 4D RAM. In the 3D case, our model must ``pay attention" to regions for each frame $\tau$. However, in the 4D case, our model does not have this requirement since $\tau$ is a free parameter. Our model has full discretion on which frames to ``pay attention to" and can move both forward and backward in time as needed. We analyze this in Figure \ref{fig:attention_4d}. Over the course of the video,  $p(\hat{y}_t = \mathbf{y})$ varies. Not only can our model change the glimpse's spatial location in each frame, it can also change the magnitude. Although our model has no explicit notion of attention magnitude, it can indirectly mimic the concept. To reduce the magnitude of attention given to frame $k$, our model moves the glimpse center to a frame further away from $k$. Although the overal ``magnitude" of attention remains constant for each glimpse, the amount of attention given to $k$ has been reduced.

As shown in Figure \ref{fig:attention_4d}, our model begins at $\varphi_1$, ``looks at" the person's shoulder, jumps to a different frame, and continues ``staring at" the shoulder. One interpretation of this is that our model has learned to identify periodic cycles.
Interestingly, it has been shown in the biological literature that males exhibit strong rotational displacement at the shoulders while walking \cite{mather1994gender}. Our model's attention corroborates this claim. The model then jumps backward in time and attends to the feet at $\varphi_3$. This indicates that leg motions (\ie gait) potentially provide traces of identity. It is quite possible that this particular glimpse path was taken since our learned policy simply never explored other paths, but our model was trained over many epochs with different initial glimpse locations to reduce this possibility.

\begin{figure}[t]
	\centering
	\vspace{-3mm}
	\includegraphics[width=0.9\linewidth]{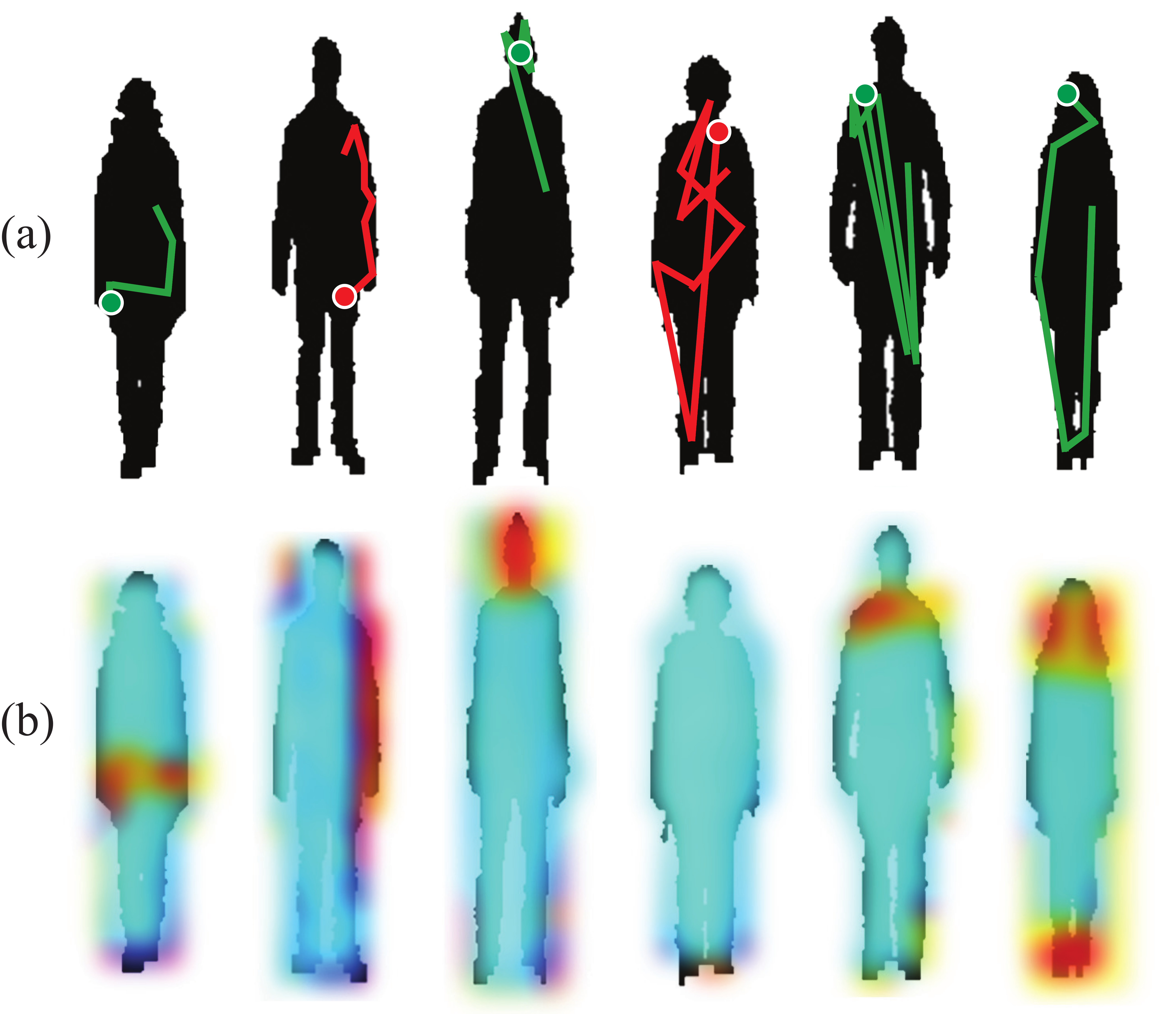}
	\caption{Two-dimensional projections of our model's 4D attention. (a) Glimpse paths. Green and red lines indicate correct and incorrect class label predictions, respectively. Circles denote the final glimpse location which led to the prediction. (b) Glimpse heatmap. Red regions denote areas on the human body frequently visited by our model. Heatmaps were smoothed with a Gaussian filter.}
	\label{fig:2d_attention}
	\vspace{-0mm}
\end{figure}

We then project the 4D attention onto a 2D image. Figure \ref{fig:2d_attention}a shows glimpse paths taken by our model. Notice how it nearly always visits a major skeleton joint. Figure \ref{fig:2d_attention}b shows an attention heatmap over all pixels. It illustrates that different regions of the body attract varying levels of attention. Our model easily identifies unique shoes or hair styles. Furthermore, it identifies the left female's hips as a discriminative region. As confirmed in the biomechanics literature \cite{cho2004gender}, females demonstrate strong lateral sway in the hip region. For some females, this alone can be the unique motion signature.

\section{Conclusion}

We introduced a recurrent attention model that identifies discriminative spatio-temporal regions for the person identification problem from depth video. Our model learns unique volumetric signatures from a high-dimensional 4D input space. Reducing the dimensionality through glimpses and an encoder allows us to train a recurrent network with a LSTM module. Evaluating our model's performance on two, three, and four dimensional inputs showed that our attention model achieves state-of-the-art performance on several person identification datasets. Visualizations of our model's attention offer new insights for future research in computer vision, biomechanics, and physiology.

\newpage
{
	\small
	\vspace{-6mm}
	\bibliographystyle{ieee}
	\bibliography{reidentification}
}

\end{document}